\documentclass[12pt]{article}
\usepackage[letterpaper, left=1in, top=1in, right=1in, bottom=1in,nohead,includefoot, verbose,ignoremp]{geometry}
\usepackage{color,charter,graphicx,natbib,amsmath,amssymb,amsfonts,amsthm,setspace,multirow,caption}
\usepackage[usenames,dvipsnames,x11names]{xcolor}
\usepackage[center]{titlesec}
\usepackage{booktabs}
\usepackage{mathtools}

\titleformat{\subsection}
  {\normalfont\large\bfseries}{\thesubsection}{1em}{}
\titlelabel{\thetitle.\quad}
\usepackage[pdftex,pagebackref=true]{hyperref}
\hypersetup{
colorlinks,
linkcolor=RoyalBlue1,
urlcolor=RoyalBlue2,
citecolor=RoyalBlue3
}
\date{}
\bibpunct{(}{)}{;}{a}{}{,}

\begin{document}

\setcounter{page}{0}\thispagestyle{empty}
\setcounter{page}{0}\thispagestyle{empty}

\title{Deep Variable-Block Chain with Adaptive Variable Selection}

\author{{Lixiang Zhang}\footnote{
	$^\dagger$ $^\ddagger$Department of Statistics, Pennsylvania State University, University Park, PA 16802, USA {\em (e-mail: \{lzz46, llin, jiali\}@psu.edu).}
},\,\, Lin Lin$^\dagger$ \& Jia Li$^\ddagger$}

\maketitle
\thispagestyle{empty}

\newpage

\newpage
\subsubsection*{Abstract}
The architectures of deep neural networks (DNN) rely heavily on the underlying grid structure of variables, for instance, the lattice of pixels in an image. For general high dimensional data with variables not associated with a grid, the multi-layer perceptron and deep belief network are often used. However, it is frequently observed that those networks do not perform competitively and they are not helpful for identifying important variables. In this paper, we propose a framework that imposes on blocks of variables a chain structure obtained by step-wise greedy search so that the DNN architecture can leverage the constructed grid. We call this new neural network Deep Variable-Block Chain (DVC). 
Because the variable blocks are used for classification in a sequential manner, we further develop the capacity of selecting variables adaptively according to a number of regions trained by a decision tree. Our experiments show that DVC outperforms other generic DNNs and other strong classifiers. Moreover, DVC can achieve high accuracy at much reduced dimensionality and sometimes reveals drastically different sets of relevant variables for different regions.

\medskip
\noindent {\em KEY WORDS:}
variable blocks, deep neural network, high-dimensional data classification, adaptive variable selection, long short-term memory  
\newpage

\section{Introduction}
Deep learning has achieved phenomenal success in a broad spectrum of predictive data analysis problems~\citep{lecun2015deep}. For a glimpse of the enormous impact, we refer to~\cite{krizhevsky2012imagenet,graves2013speech,collobert2008unified} as examples in computer vision, speech recognition, and natural language processing. Many neural network architectures have been designed for sequential and imagery data. Besides the purposes of problems in consideration, the architectures of DNN depend heavily on the underlying grid structure of the variables, e.g., pixels located on a lattice in a plane. When variables are not attributes acquired at nodes on a grid, the diversity of DNN architectures is much limited. 
Bioinformatics is an example research area in which such high-dimensional data are often handled~\citep{min2017deep}.
Several generic deep learning architectures are used in this area, including Multi-Layer Perceptron (MLP)~\citep{svozil1997introduction} and Deep Belief Networks (DBN)~\citep{hinton2006fast}. 



In this paper, we explore the idea of establishing a graph structure for the variables and in the mean time constructing a DNN on the structure. The graph structure enables us to design an architecture of less complexity than the more general neural networks such as MLP and DBN. Furthermore, the particular architecture lends itself naturally to variable selection with adaptability. Specifically, we develop a DNN called {\em Deep Variable-Block Chain (DVC)} for classification and variable selection. In addition, by exploiting the intermediate outputs of DVC, we develop an adaptive variable selection method that permits heterogeneous selection based on a decision tree. The variables are partitioned into blocks which are ordered into a chain by step-wise greedy optimization. Motivated by the chain topology of the graph, the particular architecture of DVC follows that of Long Short-Term Memory (LSTM)~\citep{gers1999learning}. However, we do not have ``time-wise'' repetitive weights as in LSTM because the chain is not a time axis and the variables along the chain are of different nature. 

Another advantage for constructing a chain and using the LSTM-like architecture is the readiness for selecting variables in an adaptive manner. To the best of our knowledge, this is an aspect unexplored and arguably not so relevant for LSTM on sequential data. As DVC outputs estimation of the class posterior probabilities through a cascade of cells, these probability values are analyzed to decide how many cells and consequently how many variables are needed for any data point. This analysis result is used to train a decision tree that determines variable selection in different regions of the data space. We would like to emphasize the difference from usual variable selection which is fixed for the whole data. The decision tree can serve solely as an adaptive assessment for the importance of variables, or it can be combined with DVC to reduce the complexity of the overall classifier.


The targeted usage of DVC is for high-dimensional data with variables not associated with a grid. If an underlying grid structure exists, we expect DNNs designed specifically for that structure to be more competitive. We thus have compared DVC with existing DNNs that are of rather generic architectures such as MLP and DBN. Experiments on several benchmark datasets show that DVC outperforms MLP and DBN in classification accuracy even though smaller sets of variables are used. The adaptive variable selection method also reveals that different numbers of variables matter in different regions of the data. This kind of insight about the variables is often valuable in the study of biological data.

The rest of the paper is organized as follows. We define notations and overview related neural networks in Section~\ref{sec:pre}. In Section~\ref{sec:chain}, we describe how the variable blocks are formed and ordered, the architecture of DVC, and its training method. The algorithm for adaptively selecting variables by a decision tree is presented in  Section~\ref{sec:tree}. Experimental results on both simulated and real data are reported in  Section~\ref{evaluation}. Finally, we conclude and discuss future work in Section~\ref{sec:conclude}.


\section{Preliminaries}
\label{sec:pre}

Denote a random vector $X$ by $(X_1,X_2,...,X_p)^{\mathrm{T}}  \in \mathbb{R}^p$ and the $i$th sample or realization of it by $(x_{i1},x_{i2},...,x_{ip})^{\mathrm{T}}  \in \mathbb{R}^p$. Moreover, the data matrix $\mathbb{X}=(x_1,x_2,...,x_p) \in \mathbb{R}^{n \times p}$, where $x_j$ is the $j$th column of $\mathbb{X}$, containing values of the $j$th variable across all sample points. We use the terms feature and variable exchangeably. A {\em variable block} is a subset of the $p$ features, e.g., $(X_1, X_3)$. Suppose we partition the $p$-dimensional random vector $X$ into $V$ variable blocks, indexed by $v= 1,2,...,V$. Let the number of variables in the $v$th block be $p_v$, a.k.a., the dimension of the $v$th variable block. We have $\sum^V_{v=1}p_v=p$. Denote the sub-vector containing variables in the $v$th variable block by $X^{(v)}$. If we reorder variables in $X$ according to the order of the variable blocks, we get the random vector $\widetilde{X}=({X^{(1)}}^{\mathrm{T}} ,{X^{(2)}}^{\mathrm{T}} ...{X^{(V)}}^{\mathrm{T}} )^{\mathrm{T}}  \in \mathbb{R}^p$. For brevity of notation, we assume without loss of generality $X^{(1)} = (X_1, X_2, ..., X_{p_1} )^{\mathrm{T}}  \in \mathbb{R}^{p_1}$ and $X^{(v)} = (X_{m_v+1}, X_{m_v+2}, ..., X_{m_v+p_v} )^{\mathrm{T}}  \in \mathbb{R}^{p_v}$, where $m_v = \sum^{v-1}_{i=1}p_i$, for $v = 2,...,V$. Then we simply have  $X=({X^{(1)}}^{\mathrm{T}} ,{X^{(2)}}^{\mathrm{T}} ...{X^{(V)}}^{\mathrm{T}} )^{\mathrm{T}}  \in \mathbb{R}^p$.

\begin{figure*}[h]
\centering
\includegraphics[width=4.5in]{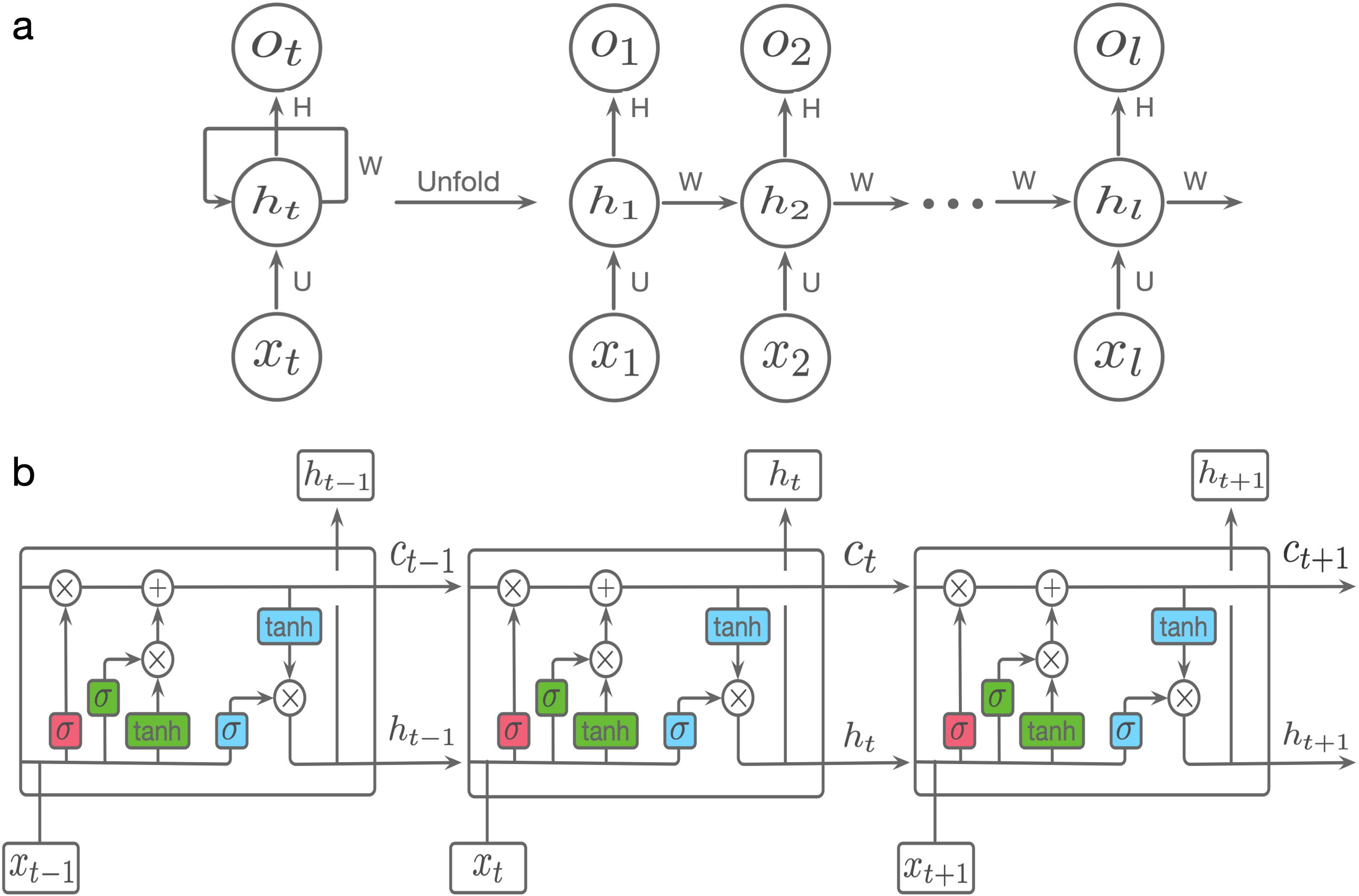}
\caption{(a) A typical RNN and the chain-like architecture if we unfold it. $W, U, H$ are weight matrices shared across the entire chain. $X$'s are input sequential data, $h$'s are hidden layers and $o$'s are outputs. (b) LSTM cell structure. The red unit at time step $t$ indicates the forget gate $f_t$, which controls the proportion of information that would the removed from previous memory cell $c_{t-1}$. The green units refer to the input gate $i_t$ and $a_t$, where $i_t$ controls the proportion of new information that would be added into current memory $c_t$ and $a_t$ generates a proposal of new information. The blue units represent the output gate $o_t$, which controls how much information would be delivered from $c_t$ to $h_t$ and influence the next cell.}
\label{lstm}
\end{figure*}

Our idea has been inspired by Recurrent Neural Network (RNN)~\citep{williams1989learning}, a type of neural networks with connections between modules forming a directed chain. Different from feedforward neural networks, the hidden states in RNN function are served as ``memory'' of the past sequence, which make RNN effective for natural language modeling~\citep{mikolov2010recurrent} and speech recognition~\citep{mikolov2011strategies}. The architecture of RNN is illustrated in Figure~\ref{lstm}(a). RNN contains a repetitive neural network module at any time position. Cascaded as a chain, these modules are ``recurrent'' since the weight matrices $W$, $U$, $H$ in each module are fixed. Let $\phi(\cdot)$ and $\psi(\cdot)$ be activation functions. In an RNN, at time $t$, the hidden unit $h_t=\phi(U X_t+W h_{t-1})$, where $X_t$ is the input data at $t$, and $h_{t-1}$, the immediate previous hidden unit, captures the effect of the past. The output unit $o_t=\psi(H h_t)$.  

RNN has the issue of vanishing gradient when the chain is long.
LSTM, a special RNN architecture, can overcome this issue by allowing gradients to flow unchanged~\citep{gers1999learning}. LSTM also contains cascaded network modules which are called in particular ``cells''. In a standard RNN, the repeating module has a relatively simple structure containing a single hyperbolic tangent hidden layer, as shown by Figure~\ref{lstm}(a). For LSTM, a typical cell is composed of a memory cell $c_t$, a hidden state $h_t$, and $3$ gates. The $3$ gates are input gate $i_t$, output gate $o_t$ and forget gate $f_t$, illustrated in Figure~\ref{lstm}(b) using a green, blue and red square respectively.

The definitions of the outputs in LSTM at every time $t$, also called updates from $t-1$ to $t$, are as follows. The notation $\odot$ means element-wise multiplication, and $\sigma$ is the sigmoid activation function with range $(0,1)$, applied to each element. Suppose $x_t\in \mathbb{R}^p$ and $h_{t-1}\in \mathbb{R}^q$, then $U_i, U_f, U_o, U_a \in \mathbb{R}^{q \times p}$, $W_i, W_f, W_o, W_a \in \mathbb{R}^{q \times q}$ and $b_i, b_f, b_o, b_a, i_t, f_t, o_t, a_t, c_t \in \mathbb{R}^q$.
\begin{align*}
i_t & = \sigma (W_{i}h_{t-1}+U_{i}x_t+b_i),\\
f_t & = \sigma (W_{f}h_{t-1}+U_{f}x_t+b_f),\\
o_t & = \sigma (W_{o}h_{t-1}+U_{o}x_t+b_o),\\
a_t & = \tanh (W_{a}h_{t-1}+U_{a}x_t+b_a),\\
c_t & = c_{t-1}\odot f_t+i_t\odot a_t, \\
h_t & = o_t\odot \tanh(c_t).
\end{align*}

As shown in Figure~\ref{lstm}(b), the memory cell $c_t$ and hidden state $h_t$ are marked at the two horizontal lines across the entire chain. They summarize the information up to $t$ and are regulated by the gates to receive new information or erase irrelevant information. The forget gate (the red unit) output is $f_t$, and $W_{f}$, $U_{f}$, $b_f$ are parameter matrices used at this gate. It receives $h_{t-1}$, the information from the last module, and the new input data $x_t$. We can view $f_t$ as a proportion to control the usage of memory $c_{t-1}$ for updating $c_{t}$ in the next cell. A higher value of  $f_t$ results in stronger influence of $c_{t-1}$ on $c_{t}$.  If the effect of the past sequence is negligible given the new input, $f_t$ approaches $0$, or figuratively, the forget gate closes. Besides past memory, the other part of $c_t$ is based on the current $i_t$ and $a_t$, which are the input gate and input proposal respectively (the two green units in Figure~\ref{lstm}(b)). The input proposal $a_t$ encodes the new information at $t$, while $i_t\in (0,1)$ controls the proportion of $a_t$ that will be added in $c_t$. The output gate $o_t \in (0,1)$ (the blue unit in Figure~\ref{lstm}(b)) controls the proportion of memory $c_t$ used to update hidden state $h_t$. At the next time $t+1$, $h_t$ will be used to deliver the encoded information up to $t$. The sequence of hidden states $h_t$ can be considered as the flow of filtered information, based on which different operations for different tasks can be defined.

The architecture of LSTM has many variations. The structure illustrated in Figure~\ref{lstm}(b) is most typical, which is flexible and relatively easy to train. To the best of our knowledge, LSTM has only been used to model sequential data. Since homogeneity is usually assumed over time, the weight and bias parameters $\theta=(W_i,U_i,b_i,W_f,U_f,b_f,W_o,U_o,b_o,W_a,U_a,b_a)$ are invariant  across the entire chain. In our work, as we construct a chain as the underlying graph for sub-vectors of variables, the LSTM architecture becomes a natural choice. Here, however, the index $t$ is not for time but for the sub-vectors $x_t$'s, which are of different meanings and possibly different dimensions. As a result, the parameters in each cell of the DNN are different. In the next section, we present the details.

\section{Deep Variable-Block Chain with Adaptive Variable Selection}
\label{sec:dvc}
\subsection{Deep Variable-Block Chain}
\label{sec:chain}

For sequential or spatial data, the variables can usually be considered as attributes of nodes on a graph. For example, pixel-wise features in an image are attributes of nodes on a regular two-dimensional grid. In another word, the indices for different variables are not symbolic, but associated with an underlying graph. To avoid confusion with the structure of a DNN, we call such underlying graph the ``field graph'', where the term ``field'' reflects the fact that the variable indices come from a physical domain. However, non-sequential or non-spatial data often contain variables not defined on a field graph. The indices for these variables can be permuted without affecting the problem, that is, the indices are actually symbolic. In such cases, some generic DNNs such as MLP are used. In MLP, variables of the input data are fully connected with the hidden layers by a weight matrix $W$ and a bias vector $b$. Let $X$ be the input data and $\phi$ be an activation function. The first hidden layer $h=\phi(W X+b)$. 
When the dimension of $X$ increases, the number of hidden units usually grows as well, resulting in quickly increasing number of parameters.
Techniques such as dropout~\citep{srivastava2014dropout} have been developed to address overfitting when the number of parameters is very large. Another practical approach is to divide the high-dimensional input matrix into several low-dimensional matrices, each used to train an MLP, and to combine the MLPs at the end. However, these techniques are limited in terms of reducing the complexity of the neural network. 

Our main idea is to simultaneously construct a field graph for variable blocks and train a DNN that exploits the graph structure. In the current work, the graph topology is a chain. We thus use a LSTM-like architecture for the DNN and call the model {\em Deep Variable-Block Chain (DVC)}. Although the variables are pre-partitioned into blocks, the chain is formed by forward step-wise greedy search and is coupled with the iterative fitting of DVC. As we will demonstrate, based on the chain structure, the variable blocks are used in a nested fashion, a fact which we exploit to develop an adaptive variable selection method (Section~\ref{sec:tree}).  

The training of DVC contains the following major steps. We will explain each step in details later.
\begin{enumerate}
    \item Partition the variables $(X_1, ..., X_p)$ into $V$ variable blocks denoted by $X^{(v)}$, $v=1, ..., V$, $X^{(v)}\in \mathbb{R}^{p_v}$.
    \item
    Let set $\mathcal{V}=\{1, ..., V\}$. At each $v=1, ..., V$, suppose the chain of variable blocks up to $v-1$ has been formed: $X^{(B_1)}$, $X^{(B_2)}$, ..., $X^{(B_{v-1})}$. Let $\mathcal{V}_{v-1}=\{B_1, ..., B_{v-1}\}$ for $v>1$ and $\mathcal{V}_0=\emptyset$. For any $i\in \mathcal{V}-\mathcal{V}_{v-1}$, assume $X^{(i)}$ is the $v$th variable block in the chain. Train a DVC using $X^{(i')}$, $i'=B_1, ..., B_{v-1}$, and $X^{(i)}$, and evaluate the classification error rate using the training data. Suppose the DVC trained by augmenting the chain with $X^{(i^{*})}$ achieves the minimum error rate. We set $B_v=i^{*}$ and repeat the process for the next $v$.
    \item
    Evaluate the cross-validation (CV) error rate for the chain at length $v$, $v=1, ..., V$, using the DVC containing variable blocks up to $v$: $X^{(B_1)}$, $X^{(B_2)}$, ..., $X^{(B_{v})}$. Suppose the minimum CV error rate is achieved when the chain contains $v^{*}$ variable blocks. We will only select variable blocks $X^{(B_v)}$, $v=1, ..., v^{*}$ for classification. 
\end{enumerate}

Remark: The third step in the algorithm essentially performs variable selection. Variables not selected in this step will not be used at all in classification. However, the adaptive variable selection method in Section~\ref{sec:tree} goes beyond this initial selection.


The motivation for using variable blocks instead of individual variables is to allow interaction among variables to be better modeled. Forming variable blocks is a way to balance model accuracy and complexity. Specifically for variable selection, grouped based selection has been much explored in statistics~\citep{yuan2006model}. We use a simple scheme to generate the variable blocks. We randomly select one seed variable and compute its correlation (or mutual information) with every other variable. Variables with the highest correlation with the seed variable are grouped with it to form a variable block. The size of the block can be decided by thresholding the correlation coefficients or by an upper bound on the number of variables permitted in one block. After one block is formed, the same process is applied to the remaining variables to form another block, so on and so forth.
By our experiments, the difference between using correlation and mutual information is negligible. We use correlation because it is faster to compute.

 

\begin{figure*}[h]
\centering
\includegraphics[width=4.5in]{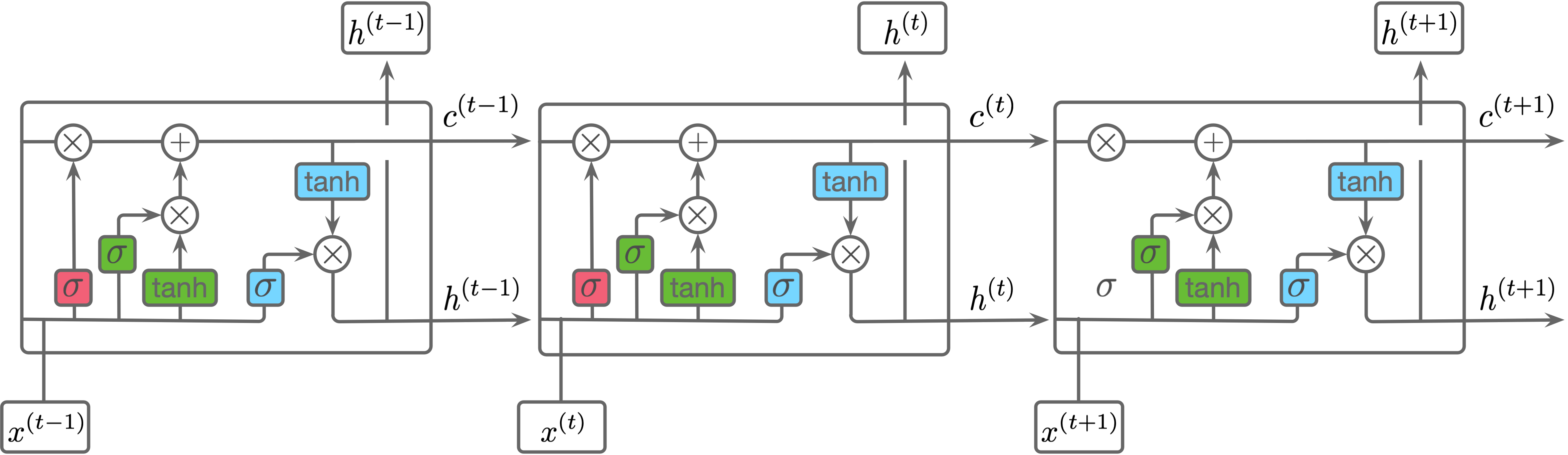}
\caption{The architecture of DVC is a cascaded sequence of cells, each taking input data from one variable block. The red unit indicates the forget gate $f^{(t)}$. The green units are the input gate $i^{(t)}$ and input proposal $a^{(t)}$. The blue square unit indicates the output gate $o^{(t)}$. The weight matrices, $W_i^{(t)}$, $W_f^{(t)}$, and $W_o^{(t)}$ are not shared across the cells.}
\label{DVC}
\end{figure*}

The architecture of DVC is essentially that of LSTM, as shown in Figure~\ref{DVC}(a). The main difference from LSTM is that the parameters in each cell are not duplicates of one set of parameters. Instead, each cell has input data of a unique variable block, and hence each cell has a unique set of parameters. Without loss of generality, suppose the input variable block to the $t$th cell is $X^{(t)}$ with realization $x^{(t)}$. Then the $t$th cell is defined as follow. 
\begin{align*}
i^{(t)} & = \sigma (W_{i}^{(t)}h^{(t-1)}+U_{i}^{(t)}x^{(t)}+b_i^{(t)}),\\
f^{(t)} & = \sigma (W_{f}^{(t)}h^{(t-1)}+U_{f}^{(t)}x^{(t)}+b_f^{(t)}),\\
o^{(t)} & = \sigma (W_{o}^{(t)}h^{(t-1)}+U_{o}^{(t)}x^{(t)}+b_o^{(t)}),\\
a^{(t)} & = \tanh (W_{a}^{(t)}h^{(t-1)}+U_{a}^{(t)}x^{(t)}+b_a^{(t)}),\\
c^{(t)} & = c^{(t-1)}\odot f^{(t)}+i^{(t)}\odot a^{(t)}, \\
h^{(t)} & = o^{(t)}\odot \tanh(c^{(t)}).
\end{align*}

Suppose $x^{(t)}\in \mathbb{R}^{p_t}$ and $h^{(t-1)}\in \mathbb{R}^q, \forall t$, then $U_i^{(t)}, U_f^{(t)}, U_o^{(t)}, U_a^{(t)} \in \mathbb{R}^{q \times p_t}$, $W_i^{(t)}, W_f^{(t)}, W_o^{(t)}$, $W_a^{(t)} \in \mathbb{R}^{q \times q}$ and $b_i^{(t)}, b_f^{(t)}, b_o^{(t)}, b_a^{(t)}, i_t^{(t)}, f_t^{(t)}, o_t^{(t)}, a_t^{(t)}, c_t^{(t)} \in \mathbb{R}^q$. In practice, we set the state size $q=\min(20, \min \{p_t, t=1,2,...,V\})$. Same as in the LSTM cell, the information is delivered across the entire chain by two horizontal lines in Figure~\ref{DVC}(a), corresponding to memory $c^{(t)}$ and hidden state $h^{(t)}$. We refer to Section~\ref{sec:pre} for detailed explanation of $f^{(t)}$, $i^{(t)}$, $a^{(t)}$, and $o^{(t)}$.
At the end of the deep chain, we add a softmax layer to output the probability of a data point belonging to each class. The softmax layer could either use the final hidden state or the entire sequence of hidden states to calculate the probabilities. Our experiments did not show considerable difference for the two options, we thus only use the final hidden state. Moreover, multiple layers can be used to define the input to each gate, but we did not observe significant improvement. 

Because the chain is formed by forward step-wise greedy search, by construction, variable blocks that contribute less to improve classification accuracy are placed at later positions in the chain. We thus perform initial selection of variable blocks by cutting off the chain at a chosen length. Suppose the initial chain contains all the $V$ variable blocks in the following order: $X^{(B_1)}$, $X^{(B_2)}$, ..., $X^{(B_V)}$. A {\em sub-chain} of length $l$ contains the first $l$ blocks: $X^{(B_1)}$, ..., $X^{(B_l)}$.
We use cross-validation to compute the classification error rate for any sub-chain of length $l$, $l=1, ..., V$.  Suppose the sub-chain of length $S$ yields the minimum CV error rate. The DVC on the sub-chain $X^{(B_1)}$, ..., $X^{(B_S)}$ is taken as the final model. We thus have performed  initial variable selection, keeping only variables in the first $S$ blocks. To distinguish from the adaptive variable selection method to be introduced in Section~\ref{sec:tree}, we call the initial selection {\em global variable selection}.

Note that because of the progressive construction of the full chain, we have fitted DVC models on all the sub-chains as a by-product when the training of the DVC using all the variable blocks is completed. The DVC models on all the sub-chains will also be used for adaptive variable selection. For clarity, we will denote the DVC trained on the sub-chain of length $l$ by $\mathcal{D}_{l}$, $l=1, ..., V$. Also note that although the sub-chains of the variable blocks are nested and $\mathcal{D}_{l}$, $l=1, ..., V$, have nested architectures, their parameters are trained separately. Thus for $l<l'$, the parameters of 
$\mathcal{D}_{l}$ are not lifted from the corresponding parameters in $\mathcal{D}_{l'}$. Finally, since the initial global variable selection decides that only the sub-chain of length $S$ ($S\leq V$) is needed, from now on, when we refer to the ``DVC on the full chain'', we mean $\mathcal{D}_S$ instead of $\mathcal{D}_V$.

We use mini-batch~\citep{li2014efficient} and Adam optimizer~\citep{kingma2014adam} to train DVC. Due to the special structure of DVC, the number of parameters in DVC is similar to that of MLP with one hidden layer. However, MLP with only one hidden layer often performs poorly, while a multi-layer MLP is prone to overfitting. 
We have observed that overfitting is less of an issue for DVC. Nevertheless, we employ L$2$ penalty for each weight matrix and the dropout technique for the input and output gates in DVC. It is found that for the experiments we conducted, these regularization techniques have no effect or at most marginal effect on performance.

\subsection{Adaptive Variable Selection by Decision Tree}
\label{sec:tree}

In this subsection, we present the adaptive variable selection method based on decision tree. The selection is adaptive in the sense that the feature space is partitioned by the decision tree and variables selected in different regions can be different. We are motivated to perform heterogeneous selection by applications in biological data analysis. For instance, bio-markers may be expected to vary by subgroups. The problem is perplexing because the subgroups are not pre-labeled or determined by some extra information but need to be discovered from the data. The decision tree is particularly appealing for finding the subgroups because its result is easy to interpret. If the subgroups were defined by complicated functions of many variables, the very purpose of selecting variables and gaining insight about their biological roles would be defied. In light of this, a shallow decision tree of a small number of leaf nodes is preferable. 

Assume the chain of variable blocks and the DVC have been trained. Recall that the variable blocks along the chain are $X^{(B_1)}$, $X^{(B_2)}$, ..., $X^{(B_S)}$, where $S\leq V$ is the number of initially selected variable blocks. For each data point $x_i$, $i=1, ..., n$, we define a so-called $\nu$-number that in a rough sense indicates the number of variable blocks needed in order to predict the class label $y_i$ correctly and with an acceptable level of ``certainty''. Let the number of classes be $K$. We now define the $\nu$-number precisely.

For brevity of notation, in the following description, we suppress the subscript index $i$ for the data point. Let the true class label for point $x$ be $y$. We apply DVC model $\mathcal{D}_l$, $l=1, ..., S$, on each sub-chain to compute the class posterior probabilities for $x$. At length $l$, let the posterior probabilities of the $K$ classes be $(b_1^{(l)}, b_2^{(l)}, ..., b_K^{(l)})$. Sort the probabilities in ascending order $(b_{(1)}^{(l)}, b_{(2)}^{(l)}, ..., b_{(K)}^{(l)})$. Define
\begin{eqnarray*}
q_l=\left\{\begin{array}{ll}
1\; , & \mbox{if}\, \, b_{y}^{(l)}=b_{(K)}^{(l)} \;\, \mbox{and}\;\, b_{y}^{(l)}-b_{(K-1)}^{(l)}\geq \epsilon \\
0 \; , & \mbox{otherwise}
\end{array}
\right . \, ,
\end{eqnarray*}
where hyper-parameter $\epsilon$ is set to $0.2/K$.

The $\nu$-number for the data point is defined by
\begin{eqnarray*}
\nu=\left\{
\begin{array}{ll}
\min\{j \in (1,2,...S), s.t. \, (q_j=1, q_{j+1}=1,..., q_{S}=1)\} & , \; \mbox{if}\, \, q_S=1\\
S+1 & , \; \mbox{if}\,\, q_S=0
\end{array}
\right . \, .
\end{eqnarray*}

Let the $\nu$-number for data points $\{x_i, i=1, ..., n\}$ be $\{\nu_i, i=1, ..., n\}$. We take $\nu_i$'s as the response variable and train a regression tree with cost-complexity pruning~\citep{breiman1984classification}. We call this decision tree the {\em variable selection tree} and denote it by $\mathcal{T}_{VS}$. 

A modification we make in $\mathcal{T}_{VS}$ from the usual regression tree is the way to decide the $\nu$-number for each leaf node of the tree. Since the variables selected at a smaller $\nu$-number are a subset of those at a larger value, we are inclined to set the $\nu$-number of a leaf node as the maximum of $\nu_i$'s belonging to this node so that all the variables needed for all the points in the leaf are included. However, this is a rather conservative criterion. Instead, we adopt a percentile strategy. Suppose the number of points with $\nu$-number equal to $l$ in a leaf node is $m_l$, $l=1, ..., S+1$. Let $m=\sum_{l=1}^{S+1}m_l$. Denote the $\nu$-number for the leaf node by $\tilde{\nu}$. Then $\tilde{\nu}=\min\{l \in (1,2,...S+1), s.t. \sum^l_{j=1}n_j \geq n\cdot\alpha\}$, where $\alpha=90\%$. 

For a region defined by a leaf node of $\mathcal{T}_{VS}$ with $\tilde{\nu}$, we select variable blocks $X^{(B_l)}$, $l=1, ..., \tilde{\nu}$, if $\tilde{\nu}\leq S$. If $\tilde{\nu}=S+1$, it indicates that a substantial portion of points in the leaf are difficult cases for classification. Since the chain has a maximum of $S$ variable blocks, when $\tilde{\nu}=S+1$, we just select all the $S$ blocks. 

We refer to the region corresponding to each leaf node of $\mathcal{T}_{VS}$ as a subgroup. $\mathcal{T}_{VS}$ provides insight into the usefulness of the variables depending on the subgroups. As aforementioned, it is desirable to have subgroups which are simply defined. In our experiments, although we did not explicitly enforce low complexity of $\mathcal{T}_{VS}$, the trees obtained are usually quite small (fewer than $4$ leaf nodes). Furthermore, as the variables are selected in groups and in a nested way, the final tree can be pruned at the choice of the user without missing useful variables. The drawback of this practice is to include variables that may have been skipped if the subgroups are more refined.

For the sole purpose of classification, we can combine $\mathcal{T}_{VS}$ with the DVC models $\mathcal{D}_l$, $l=1, ..., S$, to form an overall classifier. We call this classification method {\em DVC with Adaptive Variable Selection (DVC-AVS)}. To apply DVC-AVS, we first use $\mathcal{T}_{VS}$ to find the leaf node which a data point belongs to. Suppose the $\nu$-number of that leaf node is $\tilde{\nu}$. If $\tilde{\nu}\leq S$, $\mathcal{D}_{\tilde{\nu}}$ is used to classify the data point. If $\tilde{\nu}= S+1$, $\mathcal{D}_{S}$ is used. In contrast, the classifier DVC refers to applying $\mathcal{D}_{S}$ to any data point (no adaptive variable selection applied).


\section{Experiments}
\label{evaluation}

In this section, we use both proof-of-principle synthetic data (Section~\ref{simu}) and biomedical data (Section~\ref{real}) to demonstrate our proposed framework for classification and variable selection. For each data set, we randomly select $30\%$ of the data for testing, and the remaining $70\%$ data for training. The classification accuracy in test data, which is the proportion of correct predictions, is used to evaluate the methods. In binary classification examples, AUC (area under the ROC curve) is also used for evaluation. 
In addition, in the simulation study when we know which variables are relevant for classification, we use $F1$ score~\citep{sasaki2007truth} to evaluate the performance of variable selection. 
$F1$ score is the harmonic mean of precision and recall. Specifically,
let $\mathcal{S}$ denote the set of selected variables/features and $\mathcal{S}_r$ denote the true set of relevant variables. Then $F1={2\cdot \zeta\cdot \beta}/{(\zeta+\beta)}$, where $\zeta={|\mathcal{S}\cap \mathcal{S}_r|}/{|\mathcal{S}|}$ is the precision and $\beta={|\mathcal{S}\cap \mathcal{S}_r|}/{|\mathcal{S}_r|}$ is the recall.


\subsection{Simulation Studies}~\label{simu}

We first conduct simulation studies to evaluate DVC (with global variable selection) and DVC-AVS. We generate a sample of $1000$ points with $20$ variables from a $4$-component Gaussian mixture model. In particular, we assign equal prior probability to the $4$ components. The mean vectors for the $4$ components are $(5\cdot \mathbf{1}^T_{10}, 0\cdot \mathbf{1}^T_{10}), (10\cdot \mathbf{1}^T_{10}, 0\cdot \mathbf{1}^T_{10}), (0\cdot \mathbf{1}^T_{10}, 5\cdot \mathbf{1}^T_{10}), (0\cdot \mathbf{1}^T_{10}, 10\cdot \mathbf{1}^T_{10})$, respectively.  The covariance matrices are all set to be diagonal with diagonal elements equal to $0.25$. As the mean vectors of the $4$ components are well separated, each component is treated as one class and these $20$-dimensional variables are considered relevant variables for classification. In order to evaluate the variable selection method, we further augment each sample point by redundant variables generated independently from the standard normal distribution. These extra variables are considered irrelevant for classification. We vary the  number of total variables $p = 100,200,400$. The proportion of relevant variables is correspondingly $0.2$, $0.1$, $0.05$. 

To determine the variable block structure, we restrict that each variable block can contain only $10$ variables.  We then partition the variables according to their pairwise correlation following the procedure described in Section~\ref{sec:chain}. We observe that the $20$ relevant variables consistently form $2$ variable blocks for different $p$. Specifically, the first $10$ variables always form a block, and the next $10$ variables form another block. The two blocks of relevant variables are consistently selected as the first and second block in the chain by the algorithm; and the optimal number of variable blocks decided by global variable selection is $2$ at any $p$.

We compare the performance of DVC with $2$-hidden-layer MLP with and without the dropout technique. Neither of the two competing methods can directly select variables. The comparison of prediction accuracy is provided in Table~\ref{sim}. At any value of $p$, DVC outperforms the other two methods. 
In addition, the $F1$ score for the variables selected by DVC is consistently 1, indicating that DVC precisely identifies the relevant variables. 

\begin{table}[h]
\centering
\begin{tabular}{lcccccc}
\hline
     Accuracy             & $p=100$    & $p=200$       & $p=400$  \\\hline
DVC               & \textbf{1}               & \textbf{1}                  & \textbf{1}               \\
MLP w/o dropout & 0.974              & 0.896              & 0.843            \\
MLP w/ dropout    & 0.976            & 0.966           & 0.946           \\ \hline
\end{tabular}
\caption{Prediction accuracy for three methods under three dimensions.}
\label{sim}
\end{table}

\begin{figure*}[h]
\centering
\includegraphics[width=3.in]{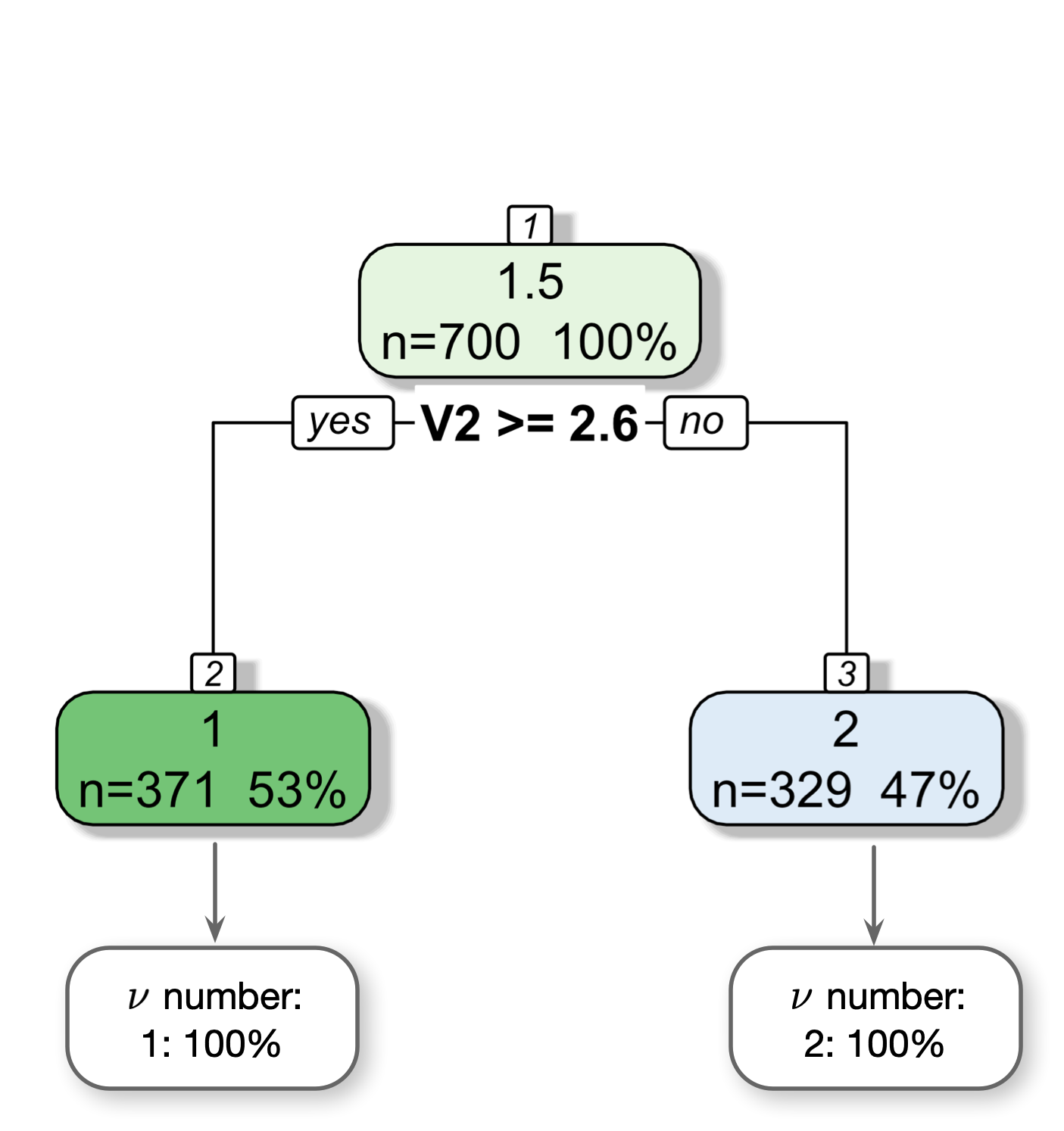}
\caption{The variable selection tree for the simulated dataset. In each node, the first number is the average $\nu$-number over points in the node, and the number above the node is the node ID. The proportions of data points with different $\nu$-numbers are shown in the white box below each leaf node. The variable used to split the node is noted beneath the node.}
\label{simtree}
\end{figure*}

The adaptive variable selection result is shown in Figure~\ref{simtree}.
The variable selection tree 
divides the space into two regions based on a single variable, in particular, the second variable in the first block (that is, the second dimension in the full data). The first subgroup, defined by the region specified by the node in dark green, contains all the points with $\nu$-number equal $1$, and the second subgroup, defined by the region specified by the node in light blue, contains points with  $\nu$-number equal $2$. By DVC-AVS,  the first subgroup can be classified using only the first variable block, and the second subgroup using the first two variable blocks. 

The above adaptive variable selection result is expected according to the design of the simulation. The first variable block can distinguish component 1 and 2, both falling in the left node with $\nu$-number equal 1; the second variable block distinguishes component 3 and 4. Due to the sequential nested selection of variables, when the $\nu$-number is $2$, the first two blocks will be selected. The classification accuracy of DVC-AVS is also $100\%$.


We further investigate the performance of DVC for variable selection in the presence of highly correlated variables. In the new set-up, the mean vectors for the $4$ components are $(5\cdot \mathbf{1}^T_{10}, 0\cdot \mathbf{1}^T_{10}), (10\cdot \mathbf{1}^T_{10}, 0\cdot \mathbf{1}^T_{10}), (-5\cdot \mathbf{1}^T_{10}, 0\cdot \mathbf{1}^T_{10}), (-10\cdot \mathbf{1}^T_{10}, 0\cdot \mathbf{1}^T_{10})$, respectively. Notice that the mean vectors for the last $10$ dimensions are identical. Conditioning on any given component, the covariance matrices for the first and second $10$ dimensions respectively are still diagonal. But we make these two variable blocks highly correlated by setting $Cov(X_i,X_{i+10})=1$, for $i=1,2,...,10$. In summary, the first 10 variables are relevant for classification, while the second 10 variables are irrelevant (but highly correlated with important variables). Similarly, we experiment with augmenting the 20 dimensions by extra irrelevant variables, the number of them ranging from $80$ to $380$. In every case, DVC only selects the first 10 variables, demonstrating that DVC is able to handle highly correlated variables in this example.

\subsection{Real Data Analysis}~\label{real}
\subsubsection{Binary Classification}
\begin{figure*}[h]
\centering
\includegraphics[width=4.in]{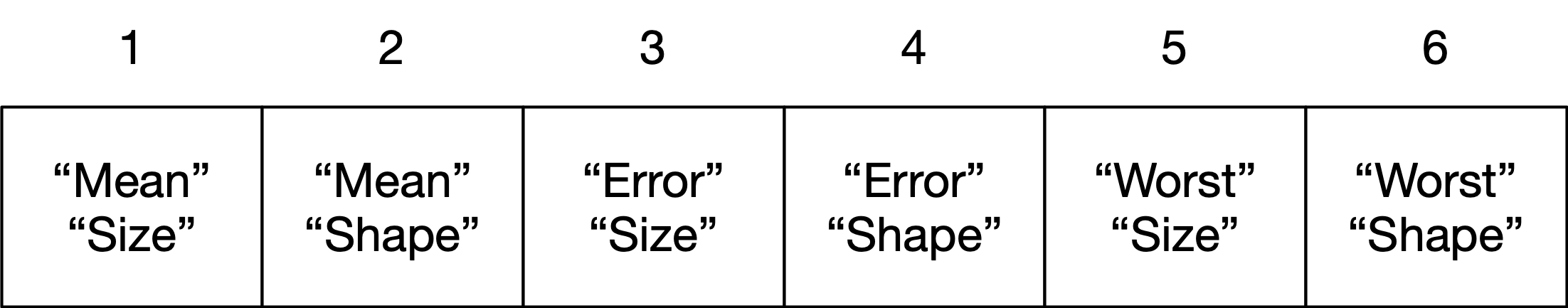}
\caption{Variable block structure of the breast cancer data. There are 6 variable blocks in total. The numbers on the first line are the variable block labels. For example, the variable block labeled as 1 contains the respective means of 5 size-related features.}
\label{wisc}
\end{figure*}

In this section, we use the Wisconsin breast cancer dataset downloaded from UCI Machine Learning Repository~\citep{wisc1995breast} to compare the performance of DVC with MLP and DBN, and two other popular machine learning methods: Support Vector Machine (SVM)~\citep{suykens1999least} with radial basis function kernel and Random Forest (RF)~\citep{breiman2001random}. The dataset contains $569$ samples with binary labels: malignant (62.7\%) and benign (37.3\%). Features are computed from digitized images of a fine needle aspirate, capturing different characteristics of the cell nuclei in the images. There are $10$ real-valued features computed for each cell nucleus per image. 
The mean, standard error, and ``worst" or largest of these  features were computed for each image, resulting in total $30$ features. Therefore, the features naturally form different groups according to two factors: the type of summary statistics (mean, error, worst), the morphological phenotypes (shape and size). Accordingly, we divide the features into $6$ variable blocks, and each variable block contains $5$ features. The description of the 6 variable blocks are shown in Figure~\ref{wisc}.


DVC determines that the optimal variable block chain is $(3,1,6,4,2)$. Since the third variable block is first chosen, this suggests that the standard errors of features related to the sizes of the cell nuclei are most effective at distinguishing the two classes if a single variable block is to be used. The test accuracy measures of DVC and several base-line methods are provided in Table~\ref{toy}. DVC achieves the highest prediction accuracy and AUC. 

\begin{table}[h]
\centering
\begin{tabular}{lcccccc}
\hline
         & MLP w/o dropout & MLP w/ dropout & DBN   & SVM   & RF    & DVC \\ \hline
Accuracy & 0.923           & 0.936          & 0.626 & 0.661 & 0.960 & \textbf{0.971} \\
AUC & 0.973           & 0.985          & 0.500 & 0.500 & 0.993 & \textbf{0.995} \\\hline
\end{tabular}\caption{Comparison of performance on breast cancer data among MLP without dropout, MLP with dropout, DBN, SVM, RF and DVC.}
\label{toy}
\end{table}

\begin{figure*}[h]
\centering
\includegraphics[width=2.5in]{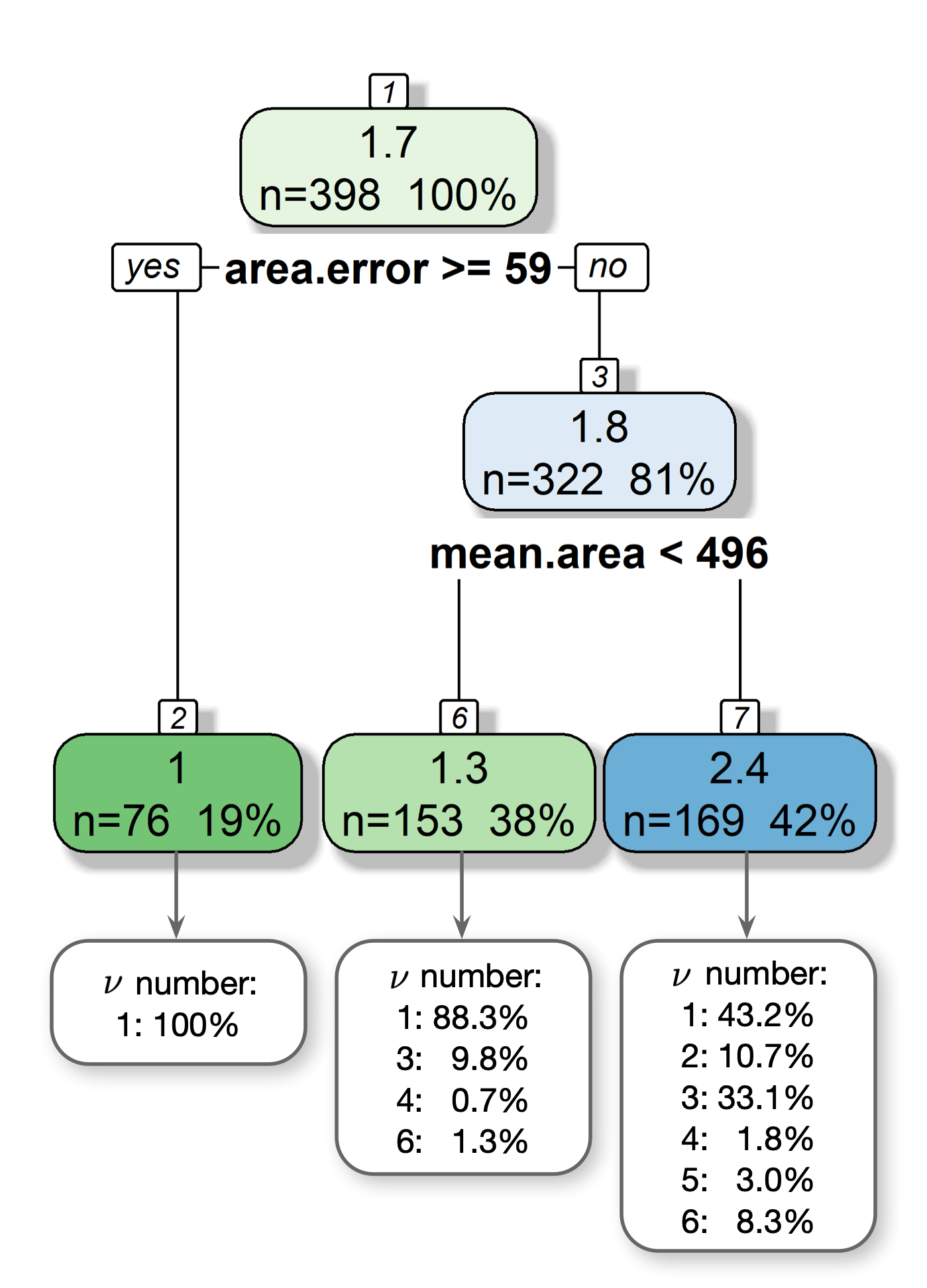}
\caption{Breast Cancer Data variable selection tree trained with the selected features and $\nu$-numbers. In each node, the first number is the average of $\nu$-numbers and the number above the node is the node ID. The proportions of data points of different $\nu$-number are shown in white box below leaf nodes $2, 6, 7$ respectively. The text below the node is the variable that used for split decision. }
\label{tree}
\end{figure*}

The variable selection tree obtained by DVC-AVS has three leaf nodes, shown in Figure~\ref{tree}. All the data points belonging to node 2 need only the first variable block ($\nu$-number = 1). The other two leaf nodes contain a mixture of  data points with different $\nu$-numbers. By the $90$th percentile strategy (Section~\ref{sec:tree}), nodes $2,6,7$ are assigned with $\nu$-number $1,3,5$ respectively. The result indicates that for different subgroups (each defined by one leaf node of the tree), different biomarkers are needed for classifying breast cancer.
We compare the accuracy of DVC and DVC-AVS for individual subgroups as well as the overall data in Table~\ref{cancer_tree}. There is no difference between the two. 

\begin{table}[h]
\centering
\begin{tabular}{lccc}
\hline
                     & Acc w/ DVC-AVS & Acc w/ DVC & Overall acc w/ DVC-AVS\\ \hline
1 $\nu$-number Group & 0.967                     &                   0.967               &               \multirow{3}{*}{0.971}       \\
3 $\nu$-number Group & 0.986                     &                 0.986                    &                   \\
5 $\nu$-number Group & 0.955                     &           0.955                          &                   \\ \hline
\end{tabular}
\caption{The comparison of group-wise and overall accuracy between DVC-AVS and DVC for Breast Cancer Data.}
\label{cancer_tree}
\end{table}

\subsubsection{Cell Type Classification}
\label{cell}

In this section, we demonstrate the use of DVC for biomarker identification for cell-type classification in single-cell data analysis. The dataset is obtained from the study of developing cerebral cortex~\citep{pollen2014low}. It consists of $301$ single cells obtained from $11$ populations: CRL-2338 epithelial (7.3\% of the total cells), CRL-2339 lymphoblastoid (5.6\%), K562 myeloid (chronic leukemia) (14\%), BJ fibroblast (from human foreskin) (12.3\%), HL60 myeloid (acute leukemia) (17.9\%), iPS pluripotent (8\%), Kera foreskin keratinocyte (13.3\%), NPC neural progenitor cells (5\%) and GW(16, 21, 21+3) fetal cortex at gestational week (16,21, 21+3 weeks) (8.6\%, 2.7\%, 5.3\%). 
After standard pre-processing steps, the training data contains $1767$ features and $210$ cells (sample size). Based on correlation, we divide the features into $V=10$, $15$, or $20$ variable blocks containing an equal number of variables. We use $V=10$ for detailed illustration of results but report accuracy for all the three cases. We have obtained similar results when mutual information is used to generate the variable blocks. The global variable selection in DVC further identifies $5$ useful variable blocks out of the total of $10$. The classification accuracy is shown in Table~\ref{pollen_acc}. DVC achieves the highest accuracy, and the results are close for different values of $V$.
\begin{table}[h]
\centering
\begin{tabular}{lcccccccc}
\hline
         & MLP  & MLP   & \multirow{2}{*}{DBN} & \multirow{2}{*}{SVM}  & \multirow{2}{*}{RF} &DVC  & DVC  &DVC \\
         & w/o dropout & w/ dropout & & & &$(10)$ & $(15)$&$(20)$\\\hline
Acc & 0.154           & 0.670          & 0.231 & 0.231 & 0.945 & \textbf{0.956} & 0.945 & \textbf{0.956} \\ \hline
\end{tabular}
\caption{Comparison of performance on Pollen's Single Cell Data among MLP without dropout, MLP with dropout, DBN, SVM, RF and DVC with 10, 15 and 20 variable blocks.}
\label{pollen_acc}
\end{table}

The variable selection tree trained for this dataset has only one leaf node with $\nu$-number set to $1$. This means that DVC-AVS decides to select only one variable block across the whole space. The classification accuracy achieved by DVC-AVS on the test data is $0.945$, only slightly lower than that obtained by DVC using $5$ variable blocks.  

\subsubsection{Breast Cancer Classification}
\label{breast}

The ability to identify specific biomarkers for different cancer types is crucial in translational research for precision medicine. We use the dataset in breast cancer study~\citep{krishnan2016piwi}, in which the objective is to investigate whether Piwi-interacting RNAs (piRNAs) are potential biomarkers for breast cancer. This dataset is of size $113$ ($102$ breast tumor tissues and $11$ normal tissues) and $676$ piRNAs.

In order to examine the variable selection ability of DVC, we normalize all $676$ piRNAs features and divide them into $V=10$, $15$, or $20$ variable blocks based on correlation. Each block has roughly equal number of variables. Again, we report detailed result for $V=10$ and the accuracy for all $V$'s. At every $V$, only the first variable block is selected by DVC, and the accuracy on testing data is given in Table~\ref{breast_bio}. For this dataset, except for SVM, every method achieves perfect classification on the test data. However, DVC only requires about $10\%$ of the variables. 

\begin{table}[h]
\centering
\begin{tabular}{lccccccc}
\hline
         & MLP  & DBN   & SVM   & RF    & DVC $(10)$ & DVC $(15)$ & DVC $(20)$ \\ \hline
Acc & \textbf{1}          & \textbf{1} & 0.971 & \textbf{1} & \textbf{1} & \textbf{1} & \textbf{1} \\ \hline
\end{tabular}
\caption{Comparison of performance on Breast Cancer piRNA Data among MLP without dropout, MLP with dropout, DBN, SVM, RF and DVC with 10, 15 and 20 variable blocks.}
\label{breast_bio}
\end{table}

\section{Conclusion and Future Work}
\label{sec:conclude}

We have developed a novel deep learning architecture, namely, DVC, by simultaneously finding an underlying chain structure for variable blocks. We target generic high-dimensional classification problems.  Instead of building architecturally homogeneous layers for all the variables like MLP, we build an individual cell for each variable block. The cells are then connected into a chain in an optimized order. This architecture enables us to use variable block cells to deliver important information and to reveal conditional interactions among variables. Moreover, it allows us to select variables effectively. Importantly, we have developed a decision-tree-based method to discover subgroups of data and select variables adaptively. Specifically, the selected variables can be different for different subgroups. 

As a limitation of our work, we note that DVC-AVS aims at generic high-dimensional data. When the data have a natural grid structure such as images, more targeted architectures are expected to have advantages. In addition, DVC-AVS selects nested sets of variables for different subgroups. We gain robustness from this restriction. However, in some applications, more flexible adaptive variable selection may be needed.


We have so far used a typical LSTM cell to model an individual variable block. A thorough investigation of all variants of LSTM can be beneficial for understanding the information flow of the variable chain. Another future direction is to consider underlying graph structures more complex than a chain.

\newpage
\bibliographystyle{plainnat}
\bibliography{dvc.bib}

\begin{thebibliography}{22}
\providecommand{\natexlab}[1]{#1}
\providecommand{\url}[1]{\texttt{#1}}
\expandafter\ifx\csname urlstyle\endcsname\relax
  \providecommand{\doi}[1]{doi: #1}\else
  \providecommand{\doi}{doi: \begingroup \urlstyle{rm}\Url}\fi

\bibitem[Breiman(2001)]{breiman2001random}
Leo Breiman.
\newblock Random forests.
\newblock \emph{Machine Learning}, 45\penalty0 (1):\penalty0 5--32, 2001.

\bibitem[Breiman et~al.(1984)Breiman, Friedman, Olshen, and
  Stone]{breiman1984classification}
Leo Breiman, Jerome Friedman, Richard Olshen, and Charles Stone.
\newblock Classification and regression trees.
\newblock 1984.

\bibitem[Collobert and Weston(2008)]{collobert2008unified}
Ronan Collobert and Jason Weston.
\newblock A unified architecture for natural language processing: Deep neural
  networks with multitask learning.
\newblock In \emph{Proceedings of the 25th International Conference on Machine
  Learning}, pages 160--167. ACM, 2008.

\bibitem[Gers et~al.(1999)Gers, Schmidhuber, and Cummins]{gers1999learning}
Felix Gers, J{\"u}rgen Schmidhuber, and Fred Cummins.
\newblock Learning to forget: Continual prediction with {LSTM}.
\newblock 1999.

\bibitem[Graves et~al.(2013)Graves, Mohamed, and Hinton]{graves2013speech}
Alex Graves, Abdel-rahman Mohamed, and Geoffrey Hinton.
\newblock Speech recognition with deep recurrent neural networks.
\newblock In \emph{2013 IEEE International Conference on Acoustics, Speech and
  Signal Processing}, pages 6645--6649. IEEE, 2013.

\bibitem[Hinton et~al.(2006)Hinton, Osindero, and Teh]{hinton2006fast}
Geoffrey Hinton, Simon Osindero, and Yee-Whye Teh.
\newblock A fast learning algorithm for deep belief nets.
\newblock \emph{Neural Computation}, 18\penalty0 (7):\penalty0 1527--1554,
  2006.

\bibitem[Kingma and Ba(2014)]{kingma2014adam}
Diederik Kingma and Jimmy Ba.
\newblock Adam: A method for stochastic optimization.
\newblock \emph{ArXiv Preprint ArXiv:1412.6980}, 2014.

\bibitem[Krishnan et~al.(2016)Krishnan, Ghosh, Graham, Mackey, Kovalchuk, and
  Damaraju]{krishnan2016piwi}
Preethi Krishnan, Sunita Ghosh, Kathryn Graham, John~R Mackey, Olga Kovalchuk,
  and Sambasivarao Damaraju.
\newblock Piwi-interacting {RNA}s and piwi genes as novel prognostic markers
  for breast cancer.
\newblock \emph{Oncotarget}, 7\penalty0 (25):\penalty0 37944, 2016.

\bibitem[Krizhevsky et~al.(2012)Krizhevsky, Sutskever, and
  Hinton]{krizhevsky2012imagenet}
Alex Krizhevsky, Ilya Sutskever, and Geoffrey Hinton.
\newblock Imagenet classification with deep convolutional neural networks.
\newblock In \emph{Advances in Neural Information Processing Systems}, pages
  1097--1105, 2012.

\bibitem[LeCun et~al.(2015)LeCun, Bengio, and Hinton]{lecun2015deep}
Yann LeCun, Yoshua Bengio, and Geoffrey Hinton.
\newblock Deep learning.
\newblock \emph{Nature}, 521\penalty0 (7553):\penalty0 436, 2015.

\bibitem[Li et~al.(2014)Li, Zhang, Chen, and Smola]{li2014efficient}
Mu~Li, Tong Zhang, Yuqiang Chen, and Alexander Smola.
\newblock Efficient mini-batch training for stochastic optimization.
\newblock In \emph{Proceedings of the 20th ACM SIGKDD International Conference
  on Knowledge Discovery and Data Mining}, pages 661--670. ACM, 2014.

\bibitem[Mikolov et~al.(2010)Mikolov, Karafi{\'a}t, Burget, {\v{C}}ernock{\`y},
  and Khudanpur]{mikolov2010recurrent}
Tom{\'a}{\v{s}} Mikolov, Martin Karafi{\'a}t, Luk{\'a}{\v{s}} Burget, Jan
  {\v{C}}ernock{\`y}, and Sanjeev Khudanpur.
\newblock Recurrent neural network based language model.
\newblock In \emph{Eleventh Annual Conference of the International Speech
  Communication Association}, 2010.

\bibitem[Mikolov et~al.(2011)Mikolov, Deoras, Povey, Burget, and
  {\v{C}}ernock{\`y}]{mikolov2011strategies}
Tom{\'a}{\v{s}} Mikolov, Anoop Deoras, Daniel Povey, Luk{\'a}{\v{s}} Burget,
  and Jan {\v{C}}ernock{\`y}.
\newblock Strategies for training large scale neural network language models.
\newblock In \emph{2011 IEEE Workshop on Automatic Speech Recognition \&
  Understanding}, pages 196--201. IEEE, 2011.

\bibitem[Min et~al.(2017)Min, Lee, and Yoon]{min2017deep}
Seonwoo Min, Byunghan Lee, and Sungroh Yoon.
\newblock Deep learning in bioinformatics.
\newblock \emph{Briefings in Bioinformatics}, 18\penalty0 (5):\penalty0
  851--869, 2017.

\bibitem[Pollen et~al.(2014)Pollen, Nowakowski, Shuga, Wang, Leyrat, Lui, Li,
  Szpankowski, Fowler, Chen, et~al.]{pollen2014low}
Alex Pollen, Tomasz Nowakowski, Joe Shuga, Xiaohui Wang, Anne Leyrat, Jan Lui,
  Nianzhen Li, Lukasz Szpankowski, Brian Fowler, Peilin Chen, et~al.
\newblock Low-coverage single-cell mrna sequencing reveals cellular
  heterogeneity and activated signaling pathways in developing cerebral cortex.
\newblock \emph{Nature Biotechnology}, 32\penalty0 (10):\penalty0 1053, 2014.

\bibitem[Sasaki et~al.(2007)]{sasaki2007truth}
Yutaka Sasaki et~al.
\newblock The truth of the f-measure.
\newblock \emph{Teach Tutor Mater}, 1\penalty0 (5):\penalty0 1--5, 2007.

\bibitem[Srivastava et~al.(2014)Srivastava, Hinton, Krizhevsky, Sutskever, and
  Salakhutdinov]{srivastava2014dropout}
Nitish Srivastava, Geoffrey Hinton, Alex Krizhevsky, Ilya Sutskever, and Ruslan
  Salakhutdinov.
\newblock Dropout: a simple way to prevent neural networks from overfitting.
\newblock \emph{The Journal of Machine Learning Research}, 15\penalty0
  (1):\penalty0 1929--1958, 2014.

\bibitem[Street(1995)]{wisc1995breast}
Nick Street.
\newblock Breast cancer wisconsin (diagnostic) data set, 1995.
\newblock URL
  \url{https://archive.ics.uci.edu/ml/datasets/Breast+Cancer+Wisconsin+(Diagnostic)}.

\bibitem[Suykens and Vandewalle(1999)]{suykens1999least}
Johan Suykens and Joos Vandewalle.
\newblock Least squares support vector machine classifiers.
\newblock \emph{Neural Processing Letters}, 9\penalty0 (3):\penalty0 293--300,
  1999.

\bibitem[Svozil et~al.(1997)Svozil, Kvasnicka, and
  Pospichal]{svozil1997introduction}
Daniel Svozil, Vladimir Kvasnicka, and Jiri Pospichal.
\newblock Introduction to multi-layer feed-forward neural networks.
\newblock \emph{Chemometrics and Intelligent Laboratory Systems}, 39\penalty0
  (1):\penalty0 43--62, 1997.

\bibitem[Williams and Zipser(1989)]{williams1989learning}
Ronald Williams and David Zipser.
\newblock A learning algorithm for continually running fully recurrent neural
  networks.
\newblock \emph{Neural Computation}, 1\penalty0 (2):\penalty0 270--280, 1989.

\bibitem[Yuan and Lin(2006)]{yuan2006model}
Ming Yuan and Yi~Lin.
\newblock Model selection and estimation in regression with grouped variables.
\newblock \emph{Journal of the Royal Statistical Society: Series B (Statistical
  Methodology)}, 68\penalty0 (1):\penalty0 49--67, 2006.

\end{thebibliography}

\end{document}